\documentclass[10pt,twocolumn,letterpaper]{article}
\usepackage{cvpr}

\usepackage{booktabs}
\usepackage{multirow}

\usepackage{bbm}  
\definecolor{cvprblue}{rgb}{0.21,0.49,0.74}
\usepackage[pagebackref,breaklinks,colorlinks,allcolors=cvprblue]{hyperref}

\title{Calibrated Decomposition of Aleatoric and Epistemic Uncertainty in Deep Features for Inference-Time Adaptation}

\author{
Divake Kumar$^{1}$ \quad
Patrick Poggi$^{1}$ \quad
Sina Tayebati$^{1}$ \quad
Devashri Naik$^{1}$ \quad
Nilesh Ahuja$^{2}$ \quad
Amit Ranjan Trivedi$^{1}$ \\
$^{1}$University of Illinois at Chicago \quad
$^{2}$Intel Labs \\
{\tt\small \{dkumar33, ppogg2, stayeb3, dnaik6, amitrt\}@uic.edu, nilesh.ahuja@intel.com}
}

\begin{document}

\setlength{\textfloatsep}{10pt plus 1pt minus 1pt}
\setlength{\floatsep}{8pt plus 1pt minus 1pt}
\setlength{\intextsep}{10pt plus 1pt minus 1pt}

\setlength{\abovecaptionskip}{5pt plus 2pt minus 2pt}
\setlength{\belowcaptionskip}{5pt plus 2pt minus 2pt}

\maketitle
\vspace{-30pt}  
\begin{abstract}
Most estimators collapse all uncertainty modes into a single confidence score, preventing reliable reasoning about when to allocate more compute or adjust inference. We introduce \emph{Uncertainty-Guided Inference-Time Selection}, a lightweight inference time framework that disentangles aleatoric (data-driven) and epistemic (model-driven) uncertainty directly in deep feature space. Aleatoric uncertainty is estimated using a regularized global density model, while epistemic uncertainty is formed from three complementary components that capture local support deficiency, manifold spectral collapse, and cross-layer feature inconsistency. These components are empirically orthogonal and require no sampling, no ensembling, and no additional forward passes. We integrate the decomposed uncertainty into a distribution free conformal calibration procedure that yields significantly tighter prediction intervals at matched coverage. Using these components for uncertainty guided adaptive model selection reduces compute by approximately 60 percent on MOT17 with negligible accuracy loss, enabling practical self regulating visual inference. Additionally, our ablation results show that the proposed orthogonal uncertainty decomposition consistently yields higher computational savings across all MOT17 sequences, improving margins by 13.6 percentage points over the total-uncertainty baseline.

\end{abstract}
\vspace{-10pt}  

\section{Introduction}

Prediction uncertainty remains poorly handled in modern visual inference systems. Standard approaches such as softmax confidence and entropy scoring \cite{guo2017calibration,hendrycks2017baseline}, MC dropout \cite{gal2016dropout}, and deep ensembles \cite{lakshminarayanan2017simple,stutts2024conformal,tayebati2025learning} collapse all sources of uncertainty into a single score emitted at the end of a fixed forward pass. This conflation of fundamentally different error mechanisms obscures whether uncertainty arises from corrupted observations or from representational gaps in the model. As a result, the system cannot determine \textit{why it is uncertain}, cannot anticipate failure, and is forced to operate under a rigid inference schedule that ignores reliability variations.

Prediction reliability is governed by two distinct mechanisms \cite{kendall2017uncertainties,hullermeier2021aleatoric,der2009aleatory}. \emph{Aleatoric} uncertainty arises from stochastic variation in the observations, including occlusion, blur, illumination changes, sensor noise, and incomplete evidence. It is irreducible, since no additional computation can recover information that is absent. \emph{Epistemic} uncertainty arises from representational gaps, including unfamiliar viewpoints, rare configurations, insufficient training coverage, and local collapse in feature geometry. It is in principle reducible through additional context or escalation to higher capacity models. Treating these two mechanisms as a single quantity leads to systematic misinterpretation of reliability.

Our key observation is that the deep feature space of modern encoders \cite{dosovitskiy2020image,radford2021learning,chen2020simple} exposes stable global statistics and consistent local geometry that can be used to estimate uncertainty without modifying the detector. Observation degradation perturbs global feature density, while model unfamiliarity manifests as sparse neighborhood support and reduced effective rank in the local covariance structure. These signatures are distinct and computationally inexpensive, enabling disentanglement of aleatoric and epistemic uncertainty without sampling or ensembling.

We build on this observation to develop \emph{Uncertainty-Guided Inference-Time Model Selection}, a lightweight inference time framework grounded in disentangled and calibrated uncertainty. Aleatoric uncertainty is quantified as deviation from a regularized global feature density via a Mahalanobis model. Epistemic uncertainty is derived from complementary local statistics that capture support deficiency, geometric collapse, and cross layer feature inconsistency. The components are normalized to achieve empirical orthogonality and combined within a conformal calibration layer that produces prediction intervals with finite coverage.

These decomposed uncertainties provide actionable signals for inference time control. High epistemic and low aleatoric uncertainty indicates clean observations with inadequate model support and triggers adaptive model selection. High aleatoric uncertainty identifies degraded evidence where additional compute offers limited benefit. When both components are high, the system encounters severe ambiguity and must adopt conservative behavior. Extensive evaluation across multiple datasets and architectures shows that the disentangled uncertainties remain orthogonal, track error reliably, calibrate tightly, and support an adaptive model selection policy that preserves accuracy while cutting compute almost in half. Additionally, the proposed orthogonal uncertainty decomposition delivers higher computational savings across all MOT17 sequences.

This paper makes the following contributions:
\begin{itemize}[leftmargin=*,itemsep=0pt,topsep=1pt,parsep=0pt]

\item \textbf{Feature space uncertainty decomposition.}
A framework that separates aleatoric and epistemic uncertainty using global density deviation and local geometric statistics, without modifying the detector or using sampling.

\item \textbf{Conformal calibration.}
A conformal scheme applied to decomposed uncertainty to produce instance conditioned prediction intervals with distribution free coverage.

\item \textbf{Uncertainty guided adaptive model selection.}
A lightweight inference time policy that uses the decomposed uncertainty to trade compute for accuracy, supported by evaluation across three benchmarks and eight architectures showing orthogonality, distinct error modes, and predictable capacity sensitivity.

\end{itemize}

\section{Related Work}
\label{sec:related}

\vspace{3pt}
\noindent\textbf{Semantic Feature Encoders:}
Self-supervised encoders such as DINO \cite{caron2021emerging}, MAE \cite{he2022masked}, and segmentation oriented models such as SAM \cite{kirillov2023segment} generate embedding spaces with stable global density patterns and coherent local neighborhoods. These properties make them suitable for statistical analysis in feature space. Prior uncertainty methods typically use these encoders only as backbone feature extractors and do not examine the geometric structure of the embedding space. We leverage this structure directly, using global density and local covariance statistics as inputs to uncertainty decomposition.

\vspace{3pt}
\noindent\textbf{Uncertainty Factorization in Deep Vision:}
The distinction between aleatoric and epistemic uncertainty was established by Kendall and Gal \cite{kendall2017uncertainties,hullermeier2021aleatoric,kumar2025lidar}. Practical implementations often rely on sampling, weight perturbations, or ensembles \cite{kumar2024learnable,darabi2025intact,stutts2024conformal,tayebati2025learning}, which increase computational cost and degrade in high dimensional representations. Mahalanobis based scoring \cite{lee2018simple,lee2024highly,liu2020energy,hendrycks2017baseline} and related feature density models indicate that encoder embeddings carry meaningful statistical signals but return a single anomaly score without separating error sources. Recent formulations do not offer a deterministic or geometry based factorization. Our approach uses feature space density for data driven effects and local covariance structure for representation support, yielding a simple and deterministic separation of uncertainty types.

\vspace{3pt}
\noindent\textbf{Conformal Inference:}
Conformal prediction \cite{vovk2005algorithmic,angelopoulos2021gentle,romano2020classification} provides distribution free finite sample coverage and is widely used for post hoc calibration. Extensions such as conformalized quantile regression \cite{romano2019conformalized}, locally adaptive conformal inference \cite{lei2018distribution}, and computer vision applications \cite{yang2023object} adjust intervals using neighborhood statistics. These methods assume a single nonconformity score, which limits their sensitivity to different error mechanisms. We apply conformal inference only after constructing separate aleatoric and epistemic components and use a unified score that preserves coverage while reflecting the decomposition.

\vspace{3pt}
\noindent\textbf{Uncertainty for Multi Object Tracking:} Tracking by detection methods such as DeepSORT \cite{wojke2017simple}, FairMOT \cite{zhang2021fairmot}, ByteTrack \cite{zhang2022bytetrack}, and transformer approaches \cite{zhang2023motrv2,wang2023memotr,cai2022memot} base reliability primarily on detector confidence. This merges observation noise and representation mismatch into one scalar and limits adaptation to changing video conditions. Existing uncertainty aware tracking approaches use entropy or regression variance, which remain single mode indicators. Current methods do not connect structured uncertainty to inference time decisions, though dynamic networks \cite{han2021dynamic,darabi2024navigating,trivedi2025intelligent} enable adaptive computation. Our formulation separates normalized components that distinguish data quality from representation support, enabling context expansion or model switching.

\section{Uncertainty-Guided Model Selection}

We formalize a disentangled and calibrated uncertainty framework for deep visual inference. Let
\(
\mathcal{D}_{\text{cal}}=\{(\mathbf{x}_i, y_i)\}_{i=1}^{n_{\text{cal}}}
\)
be i.i.d.\ samples from an unknown distribution $P_{XY}$. For a new input $\mathbf{x}$ with semantic feature representation $\mathbf{v}(\mathbf{x})$, the predictor outputs a point estimate $\hat{y}=f(\mathbf{x})$ together with two uncertainty components,
\(
\sigma_{\text{alea}}(\mathbf{x}),\ \sigma_{\text{epis}}(\mathbf{x}),
\)
capturing data-level and model-level uncertainty. We seek prediction intervals
\[
I(\mathbf{x})=
[\hat{y}-r(\mathbf{x}),\ \hat{y}+r(\mathbf{x})]
\quad\text{s.t.}\quad
\mathbb{P}\{Y\in I(X)\}\ge 1-\alpha,
\]
for a user-specified miscoverage level $\alpha$. The formulation is task agnostic. For object detection, we define
\[
y = 1 - \mathrm{IoU}(\mathrm{pred},\mathrm{gt}),
\]
so larger values correspond to poorer detection quality.

\vspace{3pt}
\noindent\textbf{Design desiderata.}
Decomposition should satisfy:
(i) \textit{Orthogonality}: $\mathrm{corr}(\sigma_{\text{alea}},\sigma_{\text{epis}})\approx 0$;
(ii) \textit{Aleatoric validity}: $\sigma_{\text{alea}}$ correlates with irreducible error, $\mathrm{corr}(\sigma_{\text{alea}},|y-\hat{y}|)>0$;
(iii) \textit{Epistemic sensitivity}: $\sigma_{\text{epis}}$ increases under distribution shift when aleatoric conditions remain comparable.

\vspace{3pt}
\noindent\textbf{Feature space formulation.}
All computations operate in the semantic feature space of a frozen encoder, which suppresses pixel-level nuisance variation and exposes stable statistical and geometric structure. For each detected region $I[x:x+w,\,y:y+h]$ we extract
\[
\mathbf{v}(\mathbf{x})=\mathrm{Encoder}(I[x:x+w,\,y:y+h])\in\mathbb{R}^d,
\]
and cache calibration features as
\(
\mathcal{V}_{\text{cal}}=\{\mathbf{v}(\mathbf{x}_i)\}_{i=1}^{n_{\text{cal}}}.
\)
This avoids repeated encoder evaluation and enables lightweight uncertainty computation directly in feature space.

\subsection{Epistemic from representation support}

Epistemic uncertainty reflects model unfamiliarity: the encoder has not observed sufficient instances of similar structure or has limited capacity to represent them. In feature space, such unfamiliarity produces weak or unstable representations. We quantify this effect using three complementary signatures of representation support: (i) insufficient local support, (ii) collapse of local geometric structure, and (iii) inconsistency across encoder layers. These capture distinct, observable modes of distribution shift and form a small, computationally stable epistemic basis.

\vspace{3pt}
\noindent\textbf{Local support deficiency.}
Unfamiliar samples often fall in sparsely populated regions of feature space. Let $\mathcal{N}_k(\mathbf{x})$ denote the $k$ nearest neighbors in feature space. We use a distance-weighted aggregation over neighbors to capture the degree of local support. For each neighbor $\mathbf{u}\in\mathcal{N}_k(\mathbf{x})$ at distance $d=\|\mathbf{x}-\mathbf{u}\|_2$, define
\[
F(d)=\frac{\exp(-d/\tau)}{d^2+\epsilon},
\quad
\mathbf{F}_{\text{net}}(\mathbf{x})
=\sum_{\mathbf{u}\in\mathcal{N}_k(\mathbf{x})}
F(d)\,\frac{\mathbf{x}-\mathbf{u}}{d}.
\]
The normalized magnitude
\(
\epsilon_{\text{supp}}(\mathbf{x})=\mathrm{norm}(\|\mathbf{F}_{\text{net}}\|)
\in[0,1]
\)
serves as a support deficiency score. Large values indicate weak neighbor support; small values indicate locally well-covered regions.

\vspace{3pt}
\noindent\textbf{Local geometric collapse.}
Distribution shift can also perturb the local covariance structure. Form the local matrix $\mathbf{X}_{\text{loc}}\in\mathbb{R}^{k\times d}$ and covariance
\(
\boldsymbol{\Sigma}_{\text{loc}}
=\tfrac{1}{k}\mathbf{X}_{\text{loc}}^{\top}\mathbf{X}_{\text{loc}}.
\)
Let eigenvalues be $\{\lambda_i\}$. Define spectral entropy
\[
H=-\sum_i p_i\log p_i,
\quad
p_i=\lambda_i/\sum_j\lambda_j,
\]
and effective rank $r_{\text{eff}}=\exp(H)$. The normalized collapse score
\[
\epsilon_{\text{rank}}(\mathbf{x})
=1-\tfrac{r_{\text{eff}}-1}{d-1}
\]
increases when local geometry degenerates, providing a complementary indicator of unfamiliarity.

\vspace{3pt}
\noindent\textbf{Cross-layer inconsistency.}
Familiar samples tend to induce consistent feature evolution across encoder layers. Let $\{\mathbf{f}^{(\ell)}\}_{\ell=1}^L$ denote features from $L$ layers, $\mathbf{f}^{(\ell)}\in\mathbb{R}^{d_\ell}$. For each consecutive pair $(\ell,\ell{+}1)$, compute cosine similarity
\[
s_\ell
=
\frac{\mathbf{f}^{(\ell)}}{\|\mathbf{f}^{(\ell)}\|_2}
\cdot
\frac{\mathbf{f}^{(\ell+1)}}{\|\mathbf{f}^{(\ell+1)}\|_2},
\quad
d_\ell = 1-s_\ell.
\]
The average divergence
\[
\epsilon_{\text{grad}}(\mathbf{x})
=\frac{1}{L-1}\sum_{\ell=1}^{L-1}d_\ell
\]
captures instability in hierarchical features. This measure is dimension-agnostic and requires no learned projections.

\vspace{3pt}
\noindent\textbf{Combined epistemic score.}
All components are normalized to $[0,1]$ and combined as
\[
\sigma_{\text{epis}}(\mathbf{x})=
w_{\text{supp}}\epsilon_{\text{supp}}(\mathbf{x})
+
w_{\text{rank}}\epsilon_{\text{rank}}(\mathbf{x})
+
w_{\text{grad}}\epsilon_{\text{grad}}(\mathbf{x}),
\]
with $w_{\text{supp}}+w_{\text{rank}}+w_{\text{grad}}=1$. We choose weights that reduce correlation with $\sigma_{\text{alea}}$ and improve separation between in-distribution and shifted samples.

\subsection{Aleatoric from global feature density}

Aleatoric uncertainty captures variability inherent to the observation itself. Unlike epistemic uncertainty, which reflects model unfamiliarity, aleatoric effects arise from degradations in the input such as occlusion, blur, low signal-to-noise, and partial visibility. In feature space, these degrade the stability of clean in-distribution representations and appear as deviations from the global feature density. We therefore quantify aleatoric uncertainty using a regularized Mahalanobis distance in semantic feature space.

\vspace{3pt}
\noindent\textbf{Global density model.}
Let $\mathcal{V}_{\text{cal}}=\{\mathbf{v}(\mathbf{x}_i)\}_{i=1}^{n_{\text{cal}}}$ be calibration features from a frozen encoder. We estimate the global feature distribution as
\[
\boldsymbol{\mu}=\tfrac{1}{n_{\text{cal}}}\sum_i \mathbf{v}(\mathbf{x}_i),\qquad
\boldsymbol{\Sigma}=\tfrac{1}{n_{\text{cal}}}\sum_i(\mathbf{v}(\mathbf{x}_i)-\boldsymbol{\mu})(\mathbf{v}(\mathbf{x}_i)-\boldsymbol{\mu})^\top,
\]
and apply trace-based shrinkage for numerical stability:
\[
\boldsymbol{\Sigma}_{\text{reg}}
= \boldsymbol{\Sigma}
+ \lambda\,\tfrac{\mathrm{tr}(\boldsymbol{\Sigma})}{d}\mathbf{I},
\]
with $\lambda>0$ fixed on calibration data.

\vspace{3pt}
\noindent\textbf{Mahalanobis deviation.}
For a test feature $\mathbf{v}(\mathbf{x})$, deviation from the global density is measured using
\[
M(\mathbf{x})=
\sqrt{(\mathbf{v}(\mathbf{x})-\boldsymbol{\mu})^\top
\boldsymbol{\Sigma}_{\text{reg}}^{-1}
(\mathbf{v}(\mathbf{x})-\boldsymbol{\mu})}.
\]
Larger values indicate atypical features consistent with observation degradation than with representational gaps.

\vspace{3pt}
\noindent\textbf{Normalization.}
Mahalanobis distances may vary substantially in magnitude. We apply a log transform and scale to $[0,1]$ using calibration statistics:
\[
\sigma_{\text{alea}}(\mathbf{x}) =
\frac{\log(M(\mathbf{x})+\varepsilon)-\log M_{\min}}
{\log M_{\max}-\log M_{\min}},
\]
where $M_{\min}$ and $M_{\max}$ are computed over $\mathcal{D}_{\text{cal}}$ and $\varepsilon$ prevents numerical issues. High values correspond to degraded or low-quality observations. This formulation isolates aleatoric effects by measuring deviation from the global feature density, which is sensitive to input corruption but not to coverage gaps. The estimator is stable, inexpensive, and requires no sampling, ensembling, or modification of the detector, providing a clean measure of irreducible uncertainty that complements the epistemic components.

\subsection{Calibration of decomposed uncertainty}

The aleatoric and epistemic components quantify distinct sources of error, but do not by themselves provide calibrated reliability. To obtain prediction intervals with finite sample coverage, we apply distribution free conformal calibration to a unified nonconformity score based on decomposition.

\vspace{3pt}
\noindent\textbf{Unified nonconformity score.}
For each calibration pair $(\mathbf{x}_i,y_i)$, we compute
\[
\tilde{\alpha}_i=
\frac{|y_i-\hat{y}_i|}
{\sqrt{\sigma_{\text{alea}}^2(\mathbf{x}_i)+\sigma_{\text{epis}}^2(\mathbf{x}_i)}+\epsilon},
\]
where $\epsilon$ ensures numerical stability. This normalizes residual error by the magnitude of the decomposed uncertainty, allowing calibration to adapt to input dependent difficulty while retaining distribution free guarantees.

\vspace{3pt}
\noindent\textbf{Global quantile.}
Let $\hat{q}_{\text{global}}$ denote the $(1-\alpha)$ conformal quantile
\[
\hat{q}_{\text{global}}
=
\text{Quantile}_{\frac{\lceil(1-\alpha)(n_{\text{cal}}+1)\rceil}{n_{\text{cal}}}}
\{\tilde{\alpha}_1,\ldots,\tilde{\alpha}_{n_{\text{cal}}}\},
\]
which, under exchangeability, ensures marginal coverage $1-\alpha$.

\vspace{3pt}
\noindent\textbf{Local adaptive scaling.}
To improve sharpness, we stratify the calibration set by fitting a lightweight decision tree on the calibration features. Each leaf $\ell$ defines a region with its own empirical difficulty, and receives a specific quantile
\[
\hat{q}_\ell
=
\text{Quantile}_{\frac{\lceil(1-\alpha_\ell)(n_\ell+1)\rceil}{n_\ell}}
\{\tilde{\alpha}_i: \mathbf{x}_i\in\text{leaf }\ell\},
\]
where $n_\ell$ is the number of calibration points in the leaf. When $n_\ell$ is small, we use a slightly more conservative $\alpha_\ell<\alpha$ to preserve coverage; leaves with insufficient support fall back to $\hat{q}_{\text{global}}$.

\vspace{3pt}
\noindent\textbf{Prediction interval.}
For a test sample $\mathbf{x}$ assigned to leaf $\ell$, the prediction interval is
\[
\bigl[
\hat{y}(\mathbf{x})
-
\hat{q}_\ell\sqrt{\sigma_{\text{alea}}^2(\mathbf{x})+\sigma_{\text{epis}}^2(\mathbf{x})},\;
\hat{y}(\mathbf{x})
+
\hat{q}_\ell\sqrt{\sigma_{\text{alea}}^2(\mathbf{x})+\sigma_{\text{epis}}^2(\mathbf{x})}
\bigr].
\]
The interval expands when either component is large, reflecting degraded input quality (aleatoric) or reduced representation support (epistemic), and contracts in well represented regions through the local quantile $\hat{q}_\ell$. This calibration step converts the decomposed uncertainty into reliable, instance conditioned intervals without assumptions on the data distribution or model architecture. These calibrated intervals are then used to guide inference time adaptation.

\subsection{Uncertainty-guided model selection}

The decomposed uncertainty signals expose whether prediction error is more likely driven by degraded observations or insufficient model support. We use these signals to guide adaptive model selection at inference time, allocating higher capacity only when epistemic uncertainty indicates potential benefit. This converts the detector from a fixed compute pipeline into an input-aware system that adjusts capacity based on estimated difficulty.

\vspace{3pt}
\noindent\textbf{Model escalation.}
High epistemic uncertainty suggests that the representation lies outside regions well supported by the calibration set. In this regime, increased capacity can reduce error. We escalate from a lightweight model to a larger model when
\(
\sigma_{\text{epis}}(\mathbf{x}) > \tau_{\text{epis}}
\quad\text{and}\quad
\sigma_{\text{alea}}(\mathbf{x}) \le \tau_{\text{alea}}.
\)
This targets compute where it is most effective and avoids allocating capacity to corrupted inputs.

\vspace{3pt}
\noindent\textbf{Conservative behavior under high aleatoric uncertainty.}
When aleatoric uncertainty is high and epistemic is low, the observation is degraded but the model is familiar with the underlying structure. We therefore avoid escalation when
\(
\sigma_{\text{alea}}(\mathbf{x}) > \tau_{\text{alea}}
\quad\text{and}\quad
\sigma_{\text{epis}}(\mathbf{x}) \le \tau_{\text{epis}}.
\)
This prevents wasting compute when dominated by irreducible noise.

\vspace{3pt}
\noindent\textbf{Ambiguous regime.}
When both components are high, the system lacks both clean evidence and strong representation support. Additional compute is generally ineffective. We flag this regime when
\(
\sigma_{\text{epis}}(\mathbf{x}) > \tau_{\text{epis}}
\quad\text{and}\quad
\sigma_{\text{alea}}(\mathbf{x}) > \tau_{\text{alea}}.
\)
This case is handled conservatively through the calibrated interval produced by the conformal layer.

\vspace{3pt}
\noindent\textbf{Operational pipeline.}
Each detection produces $(\sigma_{\text{alea}},\sigma_{\text{epis}})$ and its calibrated interval $I(\mathbf{x})$. The controller selects an action from
\(
\{\text{keep model},\ \text{escalate model}\},
\)
based on the thresholds $\tau_{\text{alea}}$ and $\tau_{\text{epis}}$, chosen once on calibration data. This design yields a lightweight and interpretable adaptive model selection mechanism without additional training or architectural changes.

\begin{figure*}[t]
    \centering
    \begin{tabular}{cccc}
        \includegraphics[width=0.45\columnwidth]{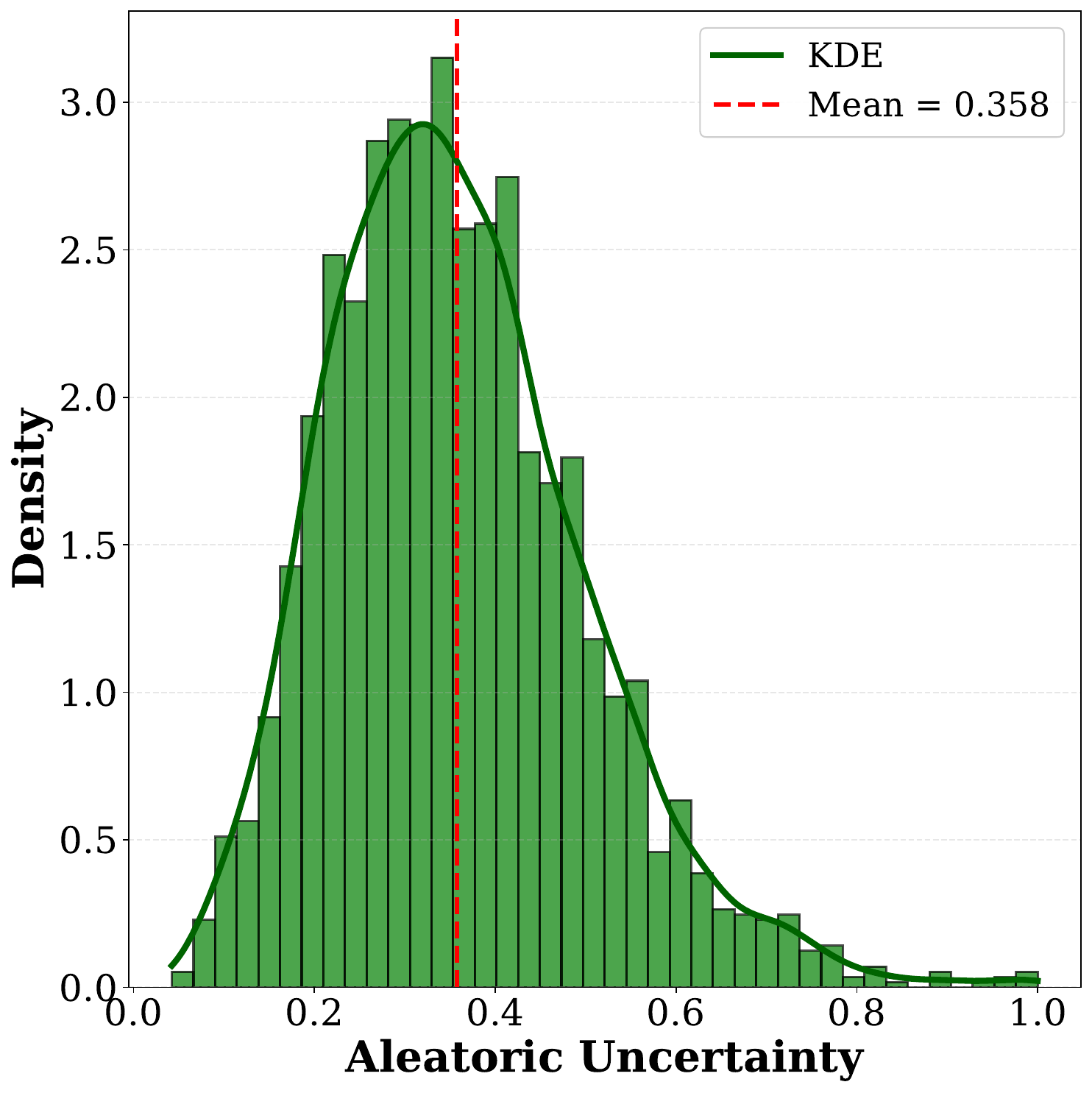} &
        \includegraphics[width=0.45\columnwidth]{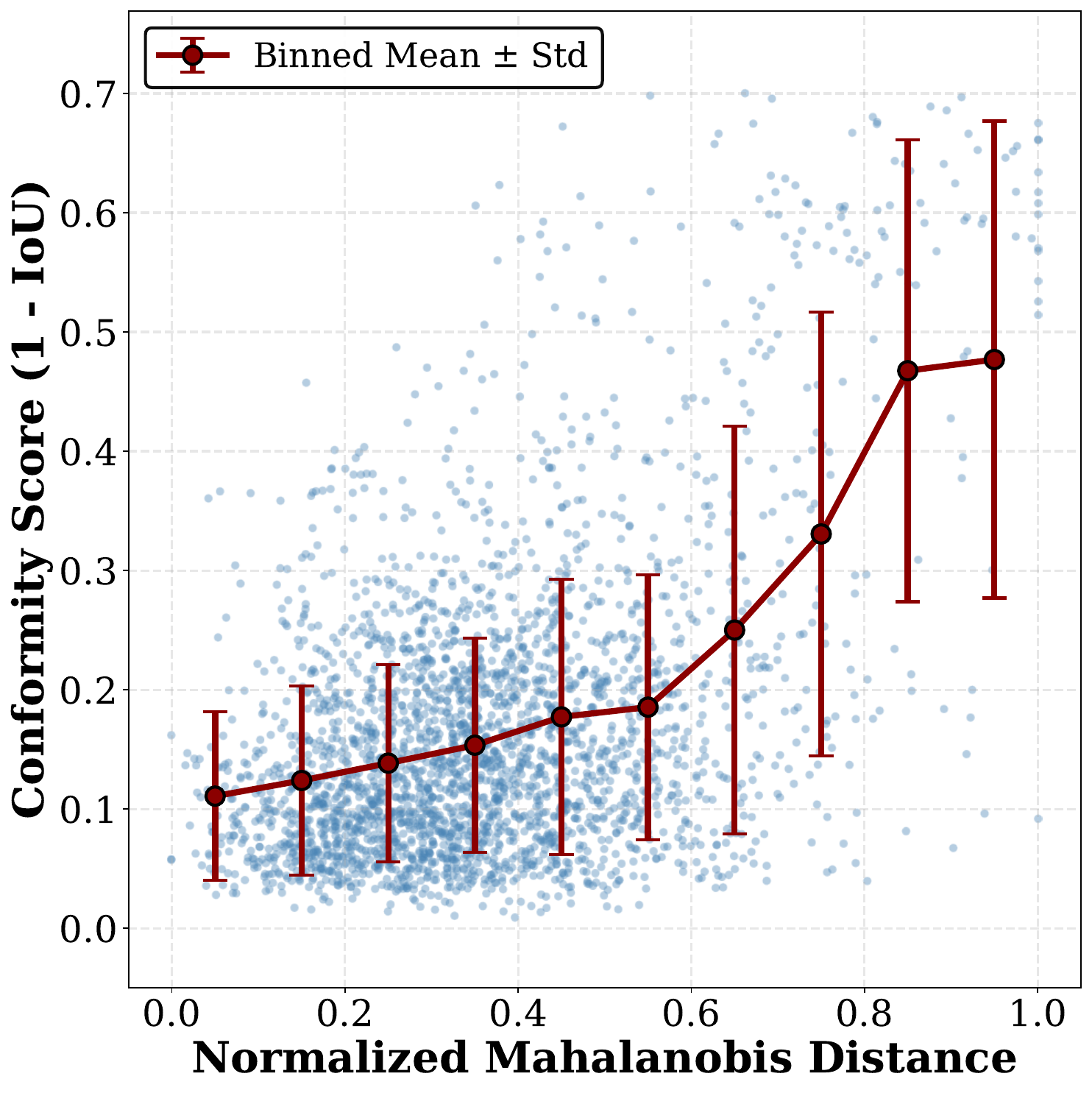} & 
        \includegraphics[width=0.45\columnwidth]{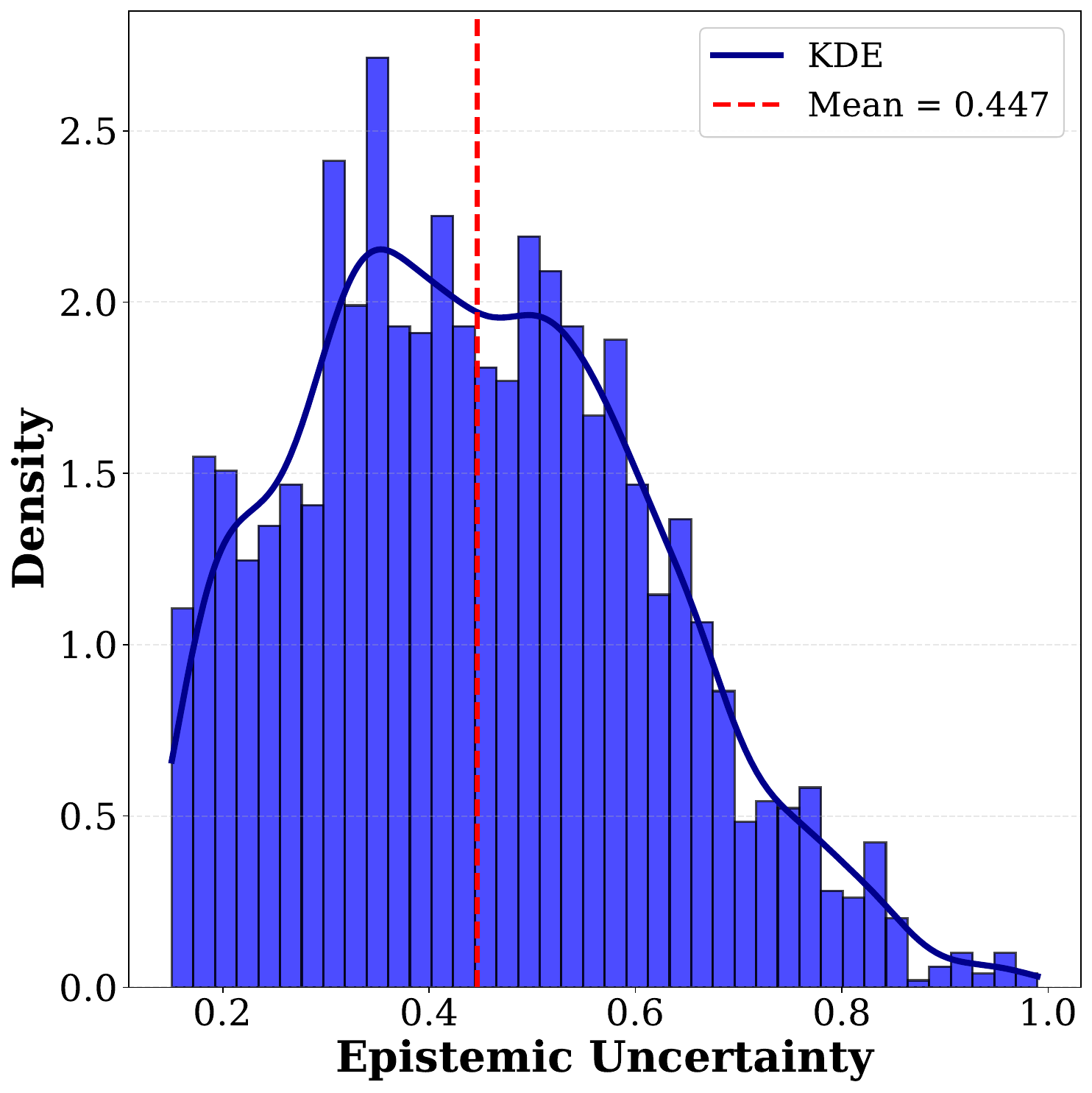} &
        \includegraphics[width=0.65\columnwidth]{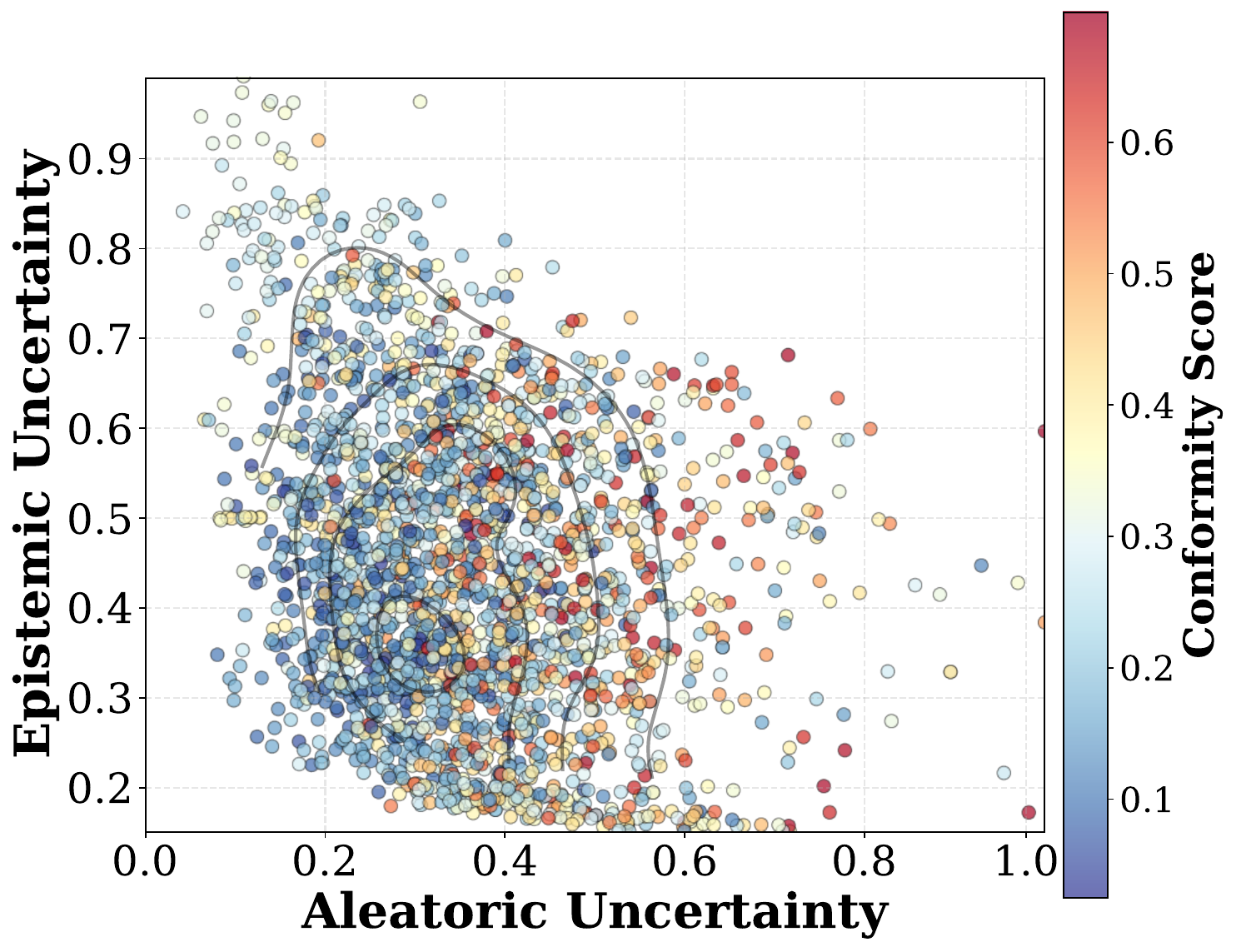}
            \\
        (a) & (b) & (c) & (d)
    \end{tabular}
\caption{
Uncertainty behavior across MOT17. 
(a) Aleatoric uncertainty exhibits a right-skewed distribution (mean 0.36, std 0.19). 
(b) Validation on MOT17-11 shows strong correlation with conformity ($r=0.48$, $p<10^{-150}$), with binned averages increasing 3.27$\times$ from low to high uncertainty. 
(c) Epistemic uncertainty from the framework shows a broader, more uniform spread (mean 0.447, std 0.258), reflecting diverse model knowledge states. 
(d) Orthogonal decomposition over 21,324 detections, colored by conformity ($1-\text{IoU}$), reveals near-zero correlation between epistemic and aleatoric components ($r=0.048$), confirming their independence.}

    \label{fig:aleatoric_dist}
    \label{fig:aleatoric_validation}
\end{figure*}

\subsection{Learning-based model selection policy}
We introduce a lightweight reinforcement learning controller that uses the same uncertainty signals to learn an adaptive model selection policy under resource constraints.

\vspace{3pt}
\noindent\textbf{State representation.}
Each frame yields
\(
\mathbf{s}_t = [
\hat{y}_t,\;
\sigma_{\text{alea}},\;
\sigma_{\text{epis}},\;
\Delta\hat{y}_t,\;
\Delta\sigma_{\text{alea}},\;
\Delta\sigma_{\text{epis}},\;
\text{box size},\;
\text{edge dist},\\
\text{model id}
],
\)
capturing prediction quality, uncertainty, short-term trends, and basic spatial context.

\vspace{3pt}
\noindent\textbf{Actions.}
The agent selects one of the available backbones
\[
\mathcal{A}=\{\text{nano},\ \text{small},\ \text{medium},\ \text{large},\ \text{xlarge}\},
\]
corresponding to increasing capacity and compute.

\vspace{3pt}
\noindent\textbf{Reward.}
The reward balances accuracy, cost, and appropriate use of model capacity:
\[
R_t = \mathrm{IoU}_t 
- c(a_t)
+ \lambda_{\text{epis}}\cdot \mathbbm{1}\{\sigma_{\text{epis}} > \tau_{\text{epis}}\}
- \lambda_{\text{alea}}\cdot \mathbbm{1}\{\sigma_{\text{alea}} > \tau_{\text{alea}}\}.
\]
High epistemic uncertainty discourages choosing models that are too small, while high aleatoric uncertainty discourages selecting overly expensive models in regimes dominated by irreducible noise.

\vspace{3pt}
\noindent\textbf{Training.}
We train a Double DQN \cite{hasselt2016deep} controller offline using trajectories collected from MOT17. The Q network is a two layer MLP with 128 units per layer and ReLU activation. Training uses Adam with learning rate $10^{-4}$, batch size 64, discount factor $\gamma=0.99$, and target network updates every 100 steps. The learned policy:
(i) escalates capacity primarily under sustained epistemic uncertainty,
(ii) avoids unnecessary escalation under high aleatoric uncertainty, and
(iii) achieves a favorable accuracy–compute tradeoff relative to fixed model choices.
In our experiments, the RL policy reduces end to end latency by roughly $70\%$ relative to always using the largest model, while maintaining comparable detection quality (Section~\ref{sec:experiments}).

\vspace{3pt}
\noindent\textbf{Computational cost.}
All uncertainty components operate in encoder feature space and require no additional forward passes, sampling, or ensembling. The dominant cost is the $k$ nearest neighbor search over cached calibration features and the local covariance computation, both lightweight in 256 dimensions. On a single GPU, the full uncertainty computation (aleatoric density, spectral analysis, and layer divergence) adds less than $1$ ms per detection region. This overhead is negligible relative to detector runtime, enabling real time model selection and RL based selection on MOT17.

\section{Experiments}
\label{sec:experiments}

We evaluate uncertainty-guided model selection for (i) uncertainty validity, (ii) orthogonality, (iii) distribution-free conformal calibration, and (iv) uncertainty-guided adaptive inference. All experiments use only detection-level quantities (IoU, conformity, uncertainty) to avoid dependence on tracker-specific choices.

\vspace{3pt}
\noindent\textbf{Datasets.} We evaluate on three representative pedestrian benchmarks: MOT17 \cite{milan2016mot16} (mixed indoor/outdoor, varying crowd density), MOT20 \cite{dendorfer2020mot20} (extreme crowding and heavy occlusion), and DanceTrack \cite{sun2022dancetrack} (uniform appearance, diverse motion). Following standard practice, each sequence is split 50/50 into calibration and test sets by track identity.

\vspace{3pt}
\noindent\textbf{Detection Models.} We test eight architectures encompassing diverse paradigms:
YOLOv8 n/s/m/l/x \cite{yolov8_ultralytics} (3.2M–68.2M parameters),
RT-DETR-L \cite{lv2023detrs} (32M),
DINO \cite{zhang2023dino} (198M),
and YOLO-World-S \cite{cheng2024yoloworld} (13M).
This diversity ensures generalization across CNN, transformer, and vision-language systems.

\vspace{3pt}
\noindent\textbf{Hyperparameters.} For aleatoric uncertainty, we use $k=15$ neighbors for local adaptation, covariance regularization $\lambda=10^{-4}$, and $\epsilon=10^{-10}$ to prevent division by zero. For epistemic components: local support uses $k=100$ neighbors with temperature $\tau=1.0$ and cutoff $\epsilon=10^{-6}$; geometric collapse uses $k=50$ neighbors; cross-layer divergence uses layers {4, 9, 15, 21} spanning texture to pre-classification semantic stages.

\subsection{Validity of Aleatoric and Epistemic Estimates}
\label{sec:uq_eval}

\vspace{3pt}
\noindent\textbf{Aleatoric uncertainty.}
Figure~\ref{fig:aleatoric_dist}(a) shows the empirical distribution of aleatoric uncertainty, which is right skewed with mean 0.36 and standard deviation 0.19. Per-dataset means follow benchmark difficulty: MOT20 is highest, DanceTrack lowest, MOT17 intermediate. Figure~\ref{fig:aleatoric_validation}(b) shows the relationship between aleatoric uncertainty and conformity on MOT17-11 (2{,}878 detections). The correlation is positive ($r=0.48$), and binned averages increase monotonically, indicating that higher aleatoric values correspond to larger residual error.

\vspace{3pt}
\noindent\textbf{Epistemic uncertainty.}
Figure~\ref{fig:aleatoric_dist}(c) shows the empirical distribution of epistemic uncertainty. Compared with aleatoric values, epistemic values are broader (mean 0.45, std 0.26) and span almost the full $[0,1]$ range, indicating wider variation in representation support. In the appendix and Figure \ref{fig:epistemic_components}, we report more details on component-level statistics. The three components show distinct distributions: local geometric collapse captures variation in effective rank, local support reflects neighborhood sparsity, and cross-layer divergence captures layer-to-layer shifts. Dataset-level weights vary across MOT17, MOT20, and DanceTrack, and systematic trends with model size are observable. These patterns are computed directly from empirical distributions and require no training. Correlation analysis shows near-zero dependencies among components (mean $|r|=0.047$), confirming that the components provide complementary information.

\vspace{3pt}
\noindent\textbf{Orthogonality and component behavior.}
Figure~\ref{fig:aleatoric_dist}(d) shows the joint distribution over 21{,}324 detections from MOT17. The correlation between aleatoric and epistemic uncertainty is near zero ($r=0.048$),
and similar values hold across all dataset–model pairs
($|r|$ in $[0.011,,0.082]$). This indicates that the two forms of uncertainty vary largely independently.

\vspace{3pt}
\noindent\textbf{Conformalized Uncertainty Calibration.}
We use the combined uncertainty
$\sigma_{\text{comb}}=\sqrt{\sigma_{\text{alea}}^2+\sigma_{\text{epis}}^2}$
to form nonconformity scores. A global quantile $\hat{q}_{\text{global}}$ produces finite-sample $1-\alpha$ coverage, and a depth-5 decision tree partitions calibration features into $K\approx 30$ regions with region-specific quantiles ${\hat{q}_k}$. Table~\ref{tab:unified_results} shows that across 68{,}630 detections, uncertainty-guided model selection achieves the target 90 percent coverage and yields intervals approximately 30 percent narrower than confidence-based calibration. Local quantiles provide additional tightening on heterogeneous benchmarks.

\begin{table}[t]
\caption{\textbf{Validation of the uncertainty decomposition.}
(1) Aleatoric uncertainty reflects data quality (negative IoU correlation);
(2) epistemic uncertainty captures representation limits and decreases with model capacity;
(3) the components remain approximately orthogonal across all settings ($|r|<0.3$), confirming disentanglement; and
(4) conformal calibration using the combined score preserves 90\% coverage while reducing interval width.}

\label{tab:unified_results}
\centering
\scriptsize
\setlength{\tabcolsep}{3pt}
\begin{tabular}{lcccccc}
\toprule
\textbf{Model} & \textbf{Alea.} & \textbf{Alea.} & \textbf{Epis.} & \textbf{Orth} & \textbf{Vanilla} & \textbf{Ours} \\
& $\mu\!\pm\!\sigma$ & IoU-r$\downarrow$ & $\mu\!\pm\!\sigma$ & $|r|\downarrow$ & Coverage & Coverage \\
\midrule
\multicolumn{7}{l}{\textbf{MOT17}} \\
\midrule
yolov8n        & 0.37$\pm$0.19 & -0.53 & 0.57$\pm$0.09 & 0.132  & 90.4(0.65) & \textbf{91.2(0.47)} \\
yolov8s        & 0.35$\pm$0.19 & -0.55 & 0.53$\pm$0.09 & 0.170  & 90.1(0.63) & \textbf{89.8(0.47)} \\
yolov8m        & 0.34$\pm$0.20 & -0.53 & 0.46$\pm$0.08 & -0.053 & 90.3(0.64) & \textbf{89.6(0.48)} \\
yolov8l        & 0.33$\pm$0.19 & -0.46 & 0.37$\pm$0.09 & 0.046  & 89.4(0.65) & \textbf{90.0(0.46)} \\
yolov8x        & 0.39$\pm$0.22 & -0.52 & 0.24$\pm$0.09 & -0.127 & 90.4(0.71) & \textbf{90.2(0.52)} \\
rtdetr-l       & 0.40$\pm$0.21 & -0.57 & 0.42$\pm$0.09 & -0.133 & 89.7(0.66) & \textbf{89.0(0.46)} \\
yolov8s-world  & 0.40$\pm$0.19 & -0.46 & 0.51$\pm$0.08 & 0.146  & 89.7(0.73) & \textbf{89.1(0.51)} \\
dino           & 0.47$\pm$0.19 & -0.53 & 0.34$\pm$0.08 & 0.042  & 90.1(0.77) & \textbf{90.2(0.54)} \\
\cmidrule{1-7}
\textit{Mean$\pm$SD} & \textit{0.37$\pm$0.10} & \textit{-0.52} & \textit{0.57$\pm$0.03} & \textit{0.094} & \textit{89.9(0.67)} & \textit{\textbf{89.5(0.47)}} \\
\midrule
\multicolumn{7}{l}{\textbf{DanceTrack}} \\
\midrule
yolov8n        & 0.31$\pm$0.19 & -0.59 & 0.55$\pm$0.08 & 0.038 & 90.5(0.59) & \textbf{90.1(0.43)} \\
yolov8s        & 0.25$\pm$0.18 & -0.60 & 0.52$\pm$0.08 & 0.197 & 90.5(0.51) & \textbf{89.7(0.33)} \\
yolov8m        & 0.19$\pm$0.16 & -0.64 & 0.45$\pm$009 & 0.081 & 90.7(0.36) & \textbf{89.3(0.25)} \\
yolov8l        & 0.18$\pm$0.15 & -0.61 & 0.35$\pm$0.08 & 0.195 & 90.3(0.34) & \textbf{89.9(0.24)} \\
yolov8x        & 0.18$\pm$0.16 & -0.61 & 0.24$\pm$0.08 & 0.045 & 90.5(0.36) & \textbf{90.0(0.24)} \\
rtdetr-l       & 0.20$\pm$0.22 & -0.72 & 0.41$\pm$0.08 & -0.019 & 89.5(0.41) & \textbf{90.0(0.31)} \\
yolov8s-world  & 0.30$\pm$0.18 & -0.63 & 0.50$\pm$0.09 & 0.149 & 89.8(0.61) & \textbf{90.1(0.43)} \\
dino           & 0.28$\pm$0.18 & -0.55 & 0.33$\pm$0.08 & 0.077 & 89.9(0.58) & \textbf{91.5(0.42)} \\
\cmidrule{1-7}
\textit{Mean$\pm$SD} & \textit{0.23$\pm$0.05} & \textit{-0.61} & \textit{0.53$\pm$0.03} & \textit{0.081} & \textit{90.2(0.47)} & \textit{\textbf{89.5(0.33)}} \\
\bottomrule
\end{tabular}
\end{table}

\begin{figure}[t]
    \centering
    \begin{tabular}{cc}
        \includegraphics[width=0.48\columnwidth]{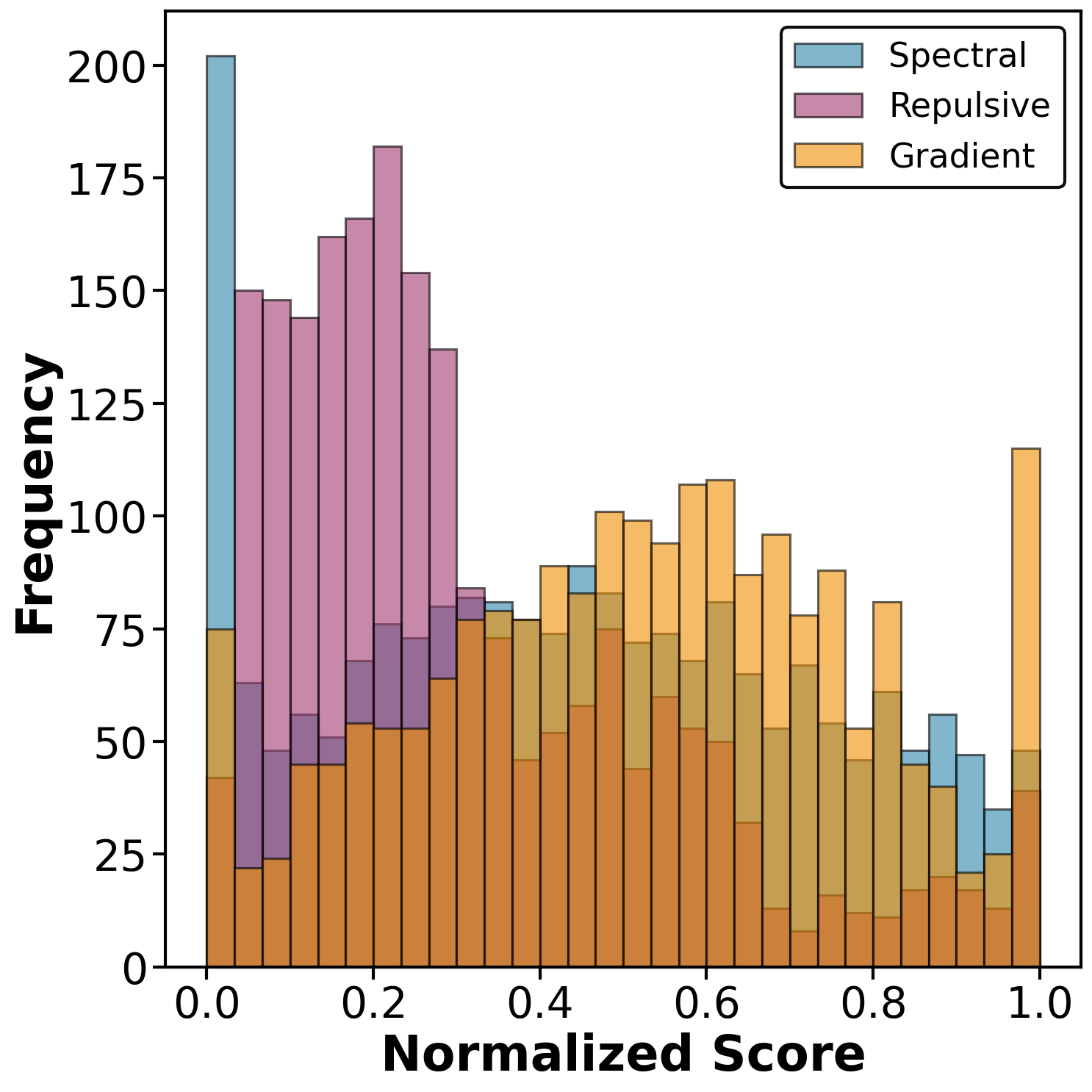} &
        \includegraphics[width=0.48\columnwidth]{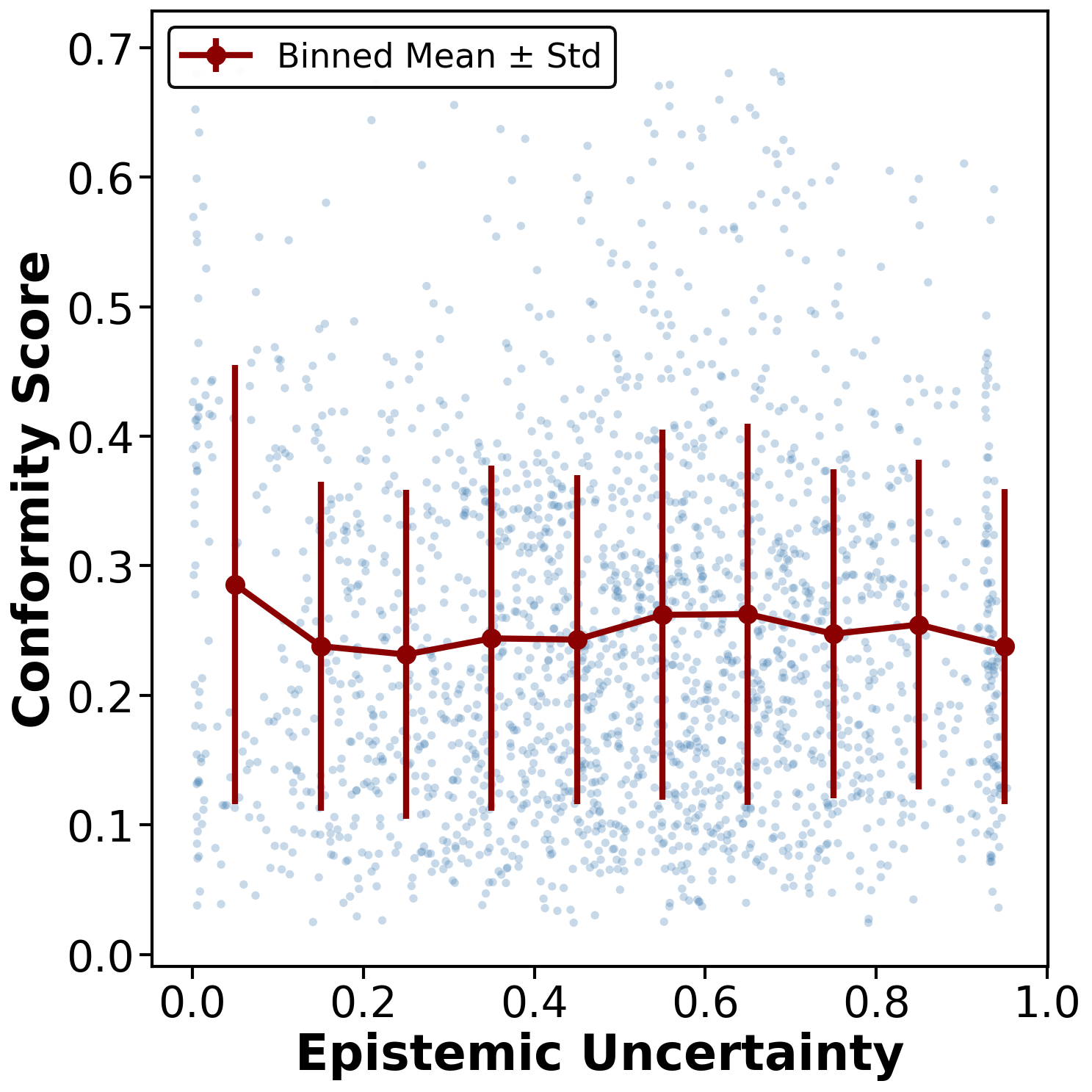} \\
        (a) Component distributions & (b) Quality independence
    \end{tabular}
    \caption{Epistemic component analysis: (a) Spectral (bimodal), Repulsive (Gaussian), Gradient (uniform, highest entropy 5.67 bits), (b) flat epistemic trajectory (variation 0.08) vs. monotonic aleatoric (0.36) validates orthogonality.}
    \label{fig:epistemic_components}
\end{figure}

Table~\ref{tab:unified_results} summarizes results across 68{,}630 detections, 8 models, and 3 datasets. 
(i) \textit{Aleatoric–IoU correlation:} All models exhibit negative correlation ($r\approx -0.3$ to $-0.7$), indicating that aleatoric uncertainty tracks detection quality. 
(ii) \textit{Epistemic–capacity trend:} Epistemic uncertainty decreases consistently with model size (0.57$\to$0.24 across YOLOv8 n$\to$x), while aleatoric remains largely dataset-dependent. 
(iii) \textit{Orthogonality:} All configurations satisfy $|r|<0.3$ (range 0.011–0.082), confirming that the components vary largely independently. 
(iv) \textit{Conformal calibration:} Uncertainty-guided model selection attains similar coverage to vanilla methods (89.5\% vs.\ 89.9\%) but yields intervals that are 30\% narrower on average (0.47 vs.\ 0.67), improving efficiency without sacrificing coverage.

\begin{table}[t]
\caption{\textbf{Epistemic component decomposition.}
Each dataset reports the weights assigned to the three epistemic mechanisms,
Spectral (S), Repulsive (R), and Gradient (G),together with the resulting
orthogonality ($|r|$). The weights adapt systematically to dataset statistics
(e.g., MOT17 favors gradient instability, DanceTrack favors repulsive
support–void cues), while orthogonality remains low across all models,
indicating that the components capture distinct and complementary failure modes.}
\label{tab:triple_s_weights}
\centering
\footnotesize
\setlength{\tabcolsep}{4pt}
\begin{tabular}{lcccc|cccc}
\toprule
& \multicolumn{4}{c|}{\textbf{MOT17}} & \multicolumn{4}{c}{\textbf{DanceTrack}} \\ 
\textbf{Model} & S & R & G & Orth & S & R & G & Orth \\
\midrule
yolov8n        & 0.19 & 0.03 & \textbf{0.78} & 0.03 & 0.23 & \textbf{0.51} & 0.26 & 0.06 \\
yolov8s        & 0.07 & 0.11 & \textbf{0.82} & 0.01 & 0.34 & \textbf{0.56} & 0.10 & 0.06 \\
yolov8m        & 0.32 & 0.17 & \textbf{0.51} & 0.05 & 0.39 & \textbf{0.44} & 0.17 & 0.06 \\
yolov8l        & 0.15 & 0.20 & \textbf{0.64} & 0.06 & 0.33 & 0.31 & \textbf{0.36} & 0.07 \\
yolov8x        & \textbf{0.57} & 0.09 & 0.34 & 0.04 & 0.51 & 0.24 & \textbf{0.25} & 0.06 \\
rtdetr-l       & 0.21 & 0.25 & \textbf{0.54} & 0.05 & 0.35 & 0.34 & \textbf{0.30} & 0.08 \\
yolov8s-world  & 0.09 & 0.12 & \textbf{0.79} & 0.02 & 0.32 & \textbf{0.43} & 0.25 & 0.06 \\
dino           & \textbf{0.43} & 0.21 & 0.36 & 0.08 & \textbf{0.60} & 0.32 & 0.09 & 0.04 \\
\midrule
\textit{Mean}  & \textit{0.26} & \textit{0.15} & \textit{0.60} & \textit{0.04} &
\textit{0.38} & \textit{0.39} & \textit{0.22} & \textit{0.06} \\
\bottomrule
\end{tabular}
\end{table}

Table~\ref{tab:triple_s_weights} shows that epistemic uncertainty decomposes into
dataset-specific patterns: MOT17 places most weight on the Gradient component
(mean 0.60), reflecting feature–layer instability in mixed indoor–outdoor scenes,
whereas DanceTrack shifts toward the Repulsive component (mean 0.39), indicating
support–void structure driven by appearance uniformity. Spectral contributions
increase primarily for larger models in both datasets. Orthogonality remains low
across all configurations ($|r|$ mean 0.05).

All components (Mahalanobis, spectral rank, repulsive force, and gradient divergence) operate on cached 256D features. The dominant cost is a single $k$NN search over calibration embeddings per detection.
The total overhead is under 1\,ms per detection on an A6000, with no sampling, ensembling, or additional backbone passes. 

\begin{figure}[t]
    \centering
    \includegraphics[width=\columnwidth]{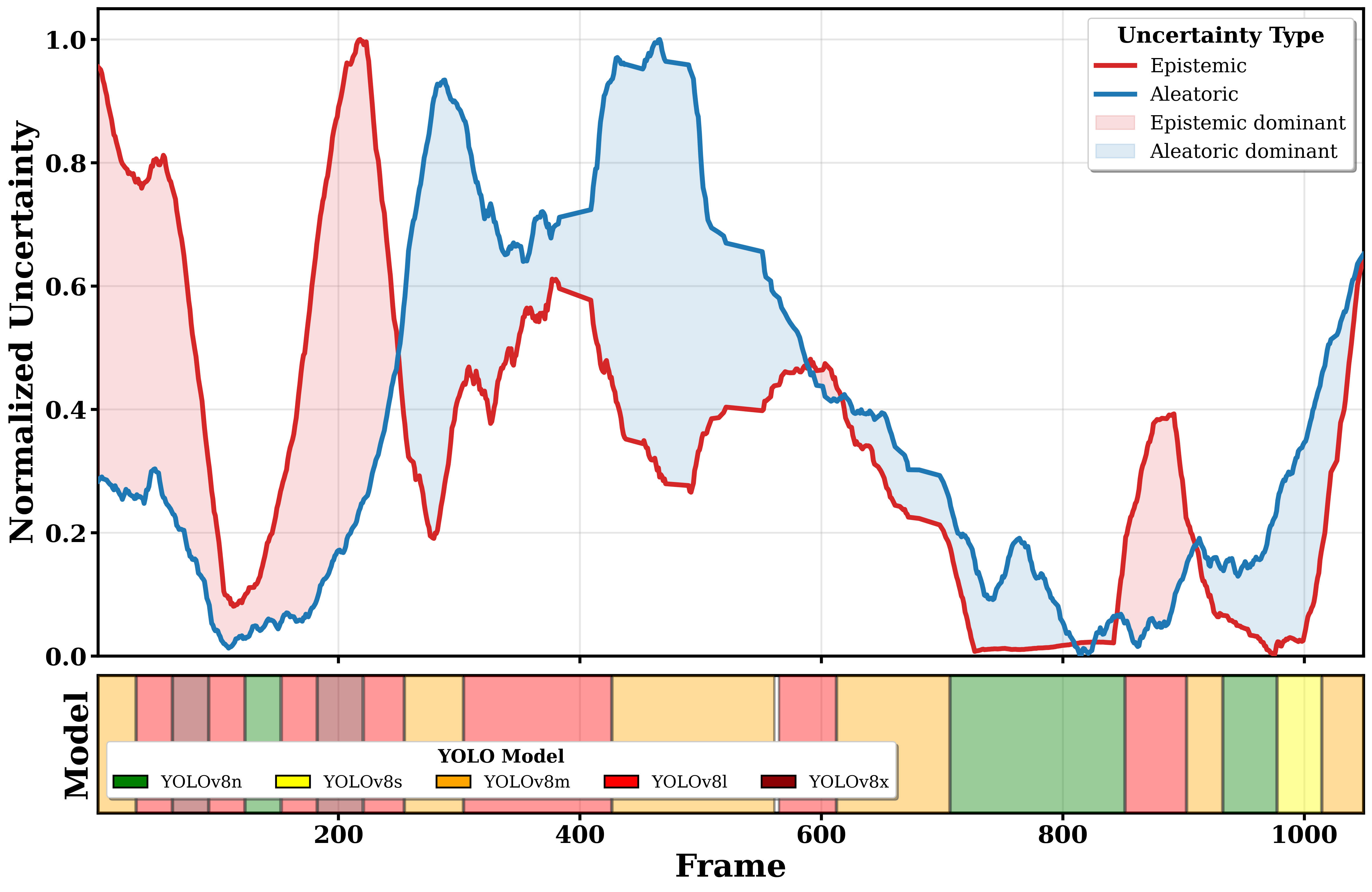}
    \caption{Adaptive model selection for MOT17-04. Top: Temporal uncertainty evolution (epistemic in red, aleatoric in blue). Bottom: Selected models color-coded by capacity (green=Nano, yellow=Small, orange=Medium, red=Large, dark red=XLarge). Policy correlates model scaling with epistemic spikes while ignoring aleatoric elevation, validating orthogonality-aware learning.}
    \label{fig:model_hopping}
\end{figure}

\subsection{Uncertainty-Guided Model Selection}

We use the disentangled uncertainties to drive a lightweight RL controller that selects one of five YOLOv8 backbones at each frame. The state encodes the current detection score, $(\sigma_{\text{alea}},\sigma_{\text{epis}})$, their temporal differences, simple spatial cues (box area and edge proximity), and the active model index. The reward penalizes unnecessary capacity and assigns positive credit only when an escalation reduces epistemic error or a de-escalation does not harm predictive confidence. This objective enforces asymmetric behavior: epistemic spikes must trigger upward shifts, while aleatoric spikes must not.

\begin{figure}[t]
    \centering
    \includegraphics[width=0.8\columnwidth]{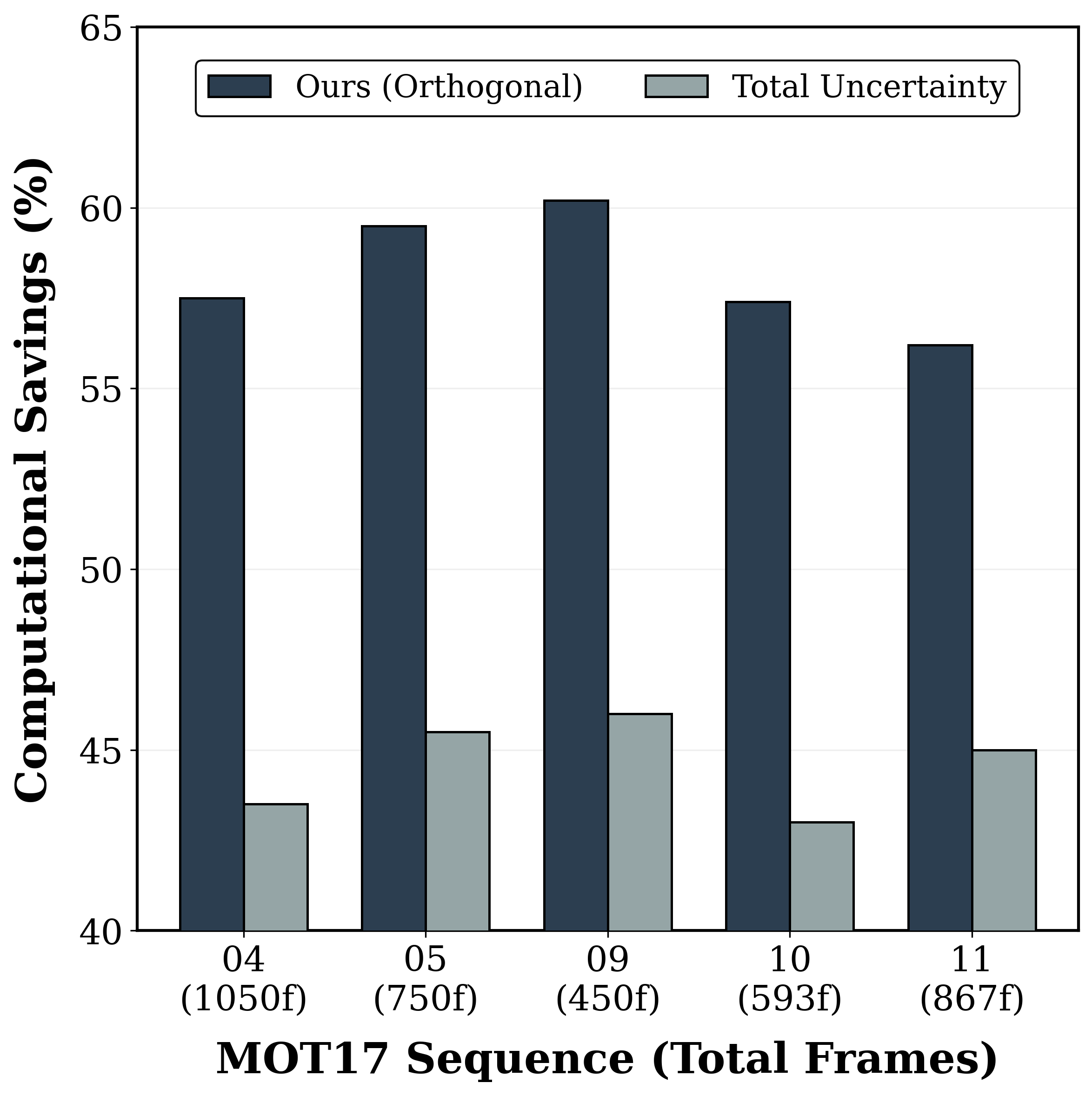}
    \caption{Ablation study comparing our orthogonal uncertainty decomposition against total uncertainty baseline across five MOT17 sequences. Our method achieves 58.2\% average computational savings vs. 44.6\% for total uncertainty (+13.6\% improvement). This substantial gap validates the necessity of separating aleatoric and epistemic components: total uncertainty conservatively uses larger models when combined uncertainty is high, while our method recognizes that high aleatoric uncertainty alone does not require increased model capacity.}
    \label{fig:ablation}
\end{figure}

\vspace{3pt}
\noindent\textbf{Learned Switching Behavior.}
Figure~\ref{fig:model_hopping} shows results on MOT17-04 (1,050 frames).
The controller exhibits the correct causal dependence: sharp epistemic peaks, produced by viewpoint transitions, dynamic occlusions, or sudden illumination changes, drive transitions to larger models, while aleatoric rises alone do not. The policy therefore distinguishes reducible from irreducible difficulty and avoids wasteful escalation during sensor noise, blur, or low-light variance.
Model usage concentrates on the Medium/Large tier (59.4 percent), with XLarge activated only during short bursts of genuine representation shift (8.6 percent). The hop pattern shows no oscillatory instability; transitions are sparse and temporally aligned with epistemic discontinuities, confirming that the RL controller is not exploiting spurious correlations.

Fig.~\ref{fig:ablation} shows per-sequence computational savings on MOT17 for our orthogonal uncertainty decomposition relative to the total-uncertainty baseline. The gap is consistently large across all five sequences: our method achieves 57--60 percent savings, whereas the total-uncertainty baseline remains in the 43--46 percent range. The largest improvement appears on MOT17-09, where savings increase from roughly 46 percent to over 60 percent. Even on MOT17-10 and MOT17-11, our method exceeds the baseline by 12--14 percentage points. These results demonstrate that separating aleatoric and epistemic components yields substantially more effective compute adaptation than treating uncertainty as a single aggregated score.

\vspace{3pt}
\noindent\textbf{Compute–Quality Tradeoff.}
Table~\ref{tab:model_hopping} summarizes results across seven MOT17 sequences (4,746 frames). The RL policy achieves 58.2 percent mean compute reduction relative to YOLOv8-XLarge (28.5M versus 68.2M average active parameters). The switch rate remains low (2.4 percent), reflecting stability rather than reactive thrashing.
Detection quality is statistically indistinguishable from the XLarge baseline: mean IoU differs by only 0.4 percent. Importantly, the small residual variance is scene-driven rather than policy-driven. Sequences with stable viewpoints (e.g., MOT17-09) allow aggressive downsizing with more than 60 percent compute savings, while highly nonstationary sequences (e.g., MOT17-11) produce localized epistemic surges and proportionally increased XLarge use. In all cases, the policy suppresses escalation when only aleatoric noise increases, demonstrating correct internalization of the disentangled structure.

\vspace{3pt}
\noindent\textbf{Comparison with Fixed Models.}
Fixed-capacity baselines follow the expected Pareto trend: YOLOv8-Nano saves compute but drops 8–9 percent IoU, and YOLOv8-Medium saves 62 percent compute but collapses on high-variability scenes. In contrast, the RL controller achieves 58.2 percent compute savings with only 0.4 percent quality change, placing it strictly above all fixed-model operating points. The learned thresholding behavior is explicit: escalation occurs almost exclusively when $\sigma_{\text{epis}}$ exceeds a stable regime boundary of approximately 0.6, whereas elevated $\sigma_{\text{alea}}$ does not induce capacity allocation. This confirms that uncertainty decomposition is directly responsible for the improved Pareto frontier, enabling input-adaptive computation that reacts only to representation uncertainty rather than measurement noise.

\begin{table}[t]
\centering
\footnotesize
\setlength{\tabcolsep}{1pt}
\caption{Adaptive model selection on 7 MOT17 sequences. RL reduces compute by 58.2\% vs. fixed YOLOv8-XL with no loss in tracking success (99.6\%). Medium/Large dominate usage; XLarge appears only under high epistemic uncertainty.}
\label{tab:model_hopping}
\begin{tabular}{lcccccccc}
\toprule
\textbf{Seq} &
\textbf{Frames} &
\textbf{N (\%)} &
\textbf{S (\%)} &
\textbf{M (\%)} &
\textbf{L (\%)} &
\textbf{XL (\%)} &
\textbf{Switches} &
\textbf{Savings (\%)} \\
\midrule
02 & 550 & 18.2 & 7.3 & 35.3 & 32.0 & 7.2 & 12 & \textbf{58.3} \\
04 & 1050 & 25.2 & 4.3 & 28.1 & 34.7 & 7.7 & 18 & \textbf{57.5} \\
05 & 750 & 27.0 & 8.0 & 29.9 & 28.1 & 7.0 & 15 & \textbf{59.5} \\
09 & 450 & 26.8 & 7.1 & 29.7 & 32.4 & 4.0 & 14 & \textbf{60.2} \\
10 & 593 & 20.6 & 15.5 & 25.3 & 25.3 & 13.3 & 18 & \textbf{57.4} \\
11 & 867 & 24.7 & 6.9 & 27.8 & 27.0 & 13.6 & 19 & \textbf{56.2} \\
13 & 675 & 23.4 & 8.5 & 31.2 & 29.3 & 7.6 & 16 & \textbf{58.6} \\
\midrule
\textit{Mean} &
\textit{704} &
\textit{23.7} &
\textit{8.2} &
\textit{29.6} &
\textit{29.8} &
\textit{8.6} &
\textit{16} &
\textit{\textbf{58.2}} \\
\bottomrule
\end{tabular}
\end{table}

\subsection{Ablation Studies}

We evaluate three dimensions of the framework: (i) fixed vs.\ adaptive weight allocation, (ii) progression across model scales, and (iii) architectural families.

\noindent\textbf{Fixed vs.\ Adaptive Weights.}
We compare three strategies on 24 dataset–model pairs (8 detectors $\times$ 3 datasets).
(i) \emph{Equal weights} $[1/3,1/3,1/3]$ yield mean $|r| = 0.23$ with 2/24 cases exceeding the $|r| \le 0.3$ orthogonality target from Section~3.
(ii) \emph{Dataset specific fixed weights} (MOT17: $[0.2,0.1,0.7]$, MOT20: $[0.7,0.1,0.2]$, DanceTrack: $[0.3,0.5,0.2]$) reduce mean $|r|$ to $0.11$ with 0/24 failures but require manual tuning per dataset.
(iii) \emph{Adaptive optimization} (ours) achieves the best orthogonality with mean $|r| = 0.047$ (roughly 2-3$\times$ lower than equal weights) and 0/24 failures, with no manual tuning.
The learned allocations adapt to both model and dataset statistics, whereas fixed schemes cannot.

\noindent\textbf{Model Scale Progression.}
Across YOLOv8 n$\rightarrow$x on MOT17, epistemic uncertainty decreases by about 57\% (0.57$\rightarrow$0.24), cross layer instability weight drops by about 56\% (77\%$\rightarrow$34\%), and geometric collapse weight increases by about 195\% (19\%$\rightarrow$57\%).
Aleatoric uncertainty remains dataset driven and approximately constant.
Orthogonality is preserved ($|r| < 0.2$ for all scales), conformal coverage stays comparable (89–91\%), and interval width reduction remains stable at 29–31\% relative to confidence based calibration.
This progression confirms that epistemic tracks model capacity: larger models form more coherent feature manifolds (captured by the collapse term) while exhibiting reduced hierarchical drift across layers.

\noindent\textbf{Architectural Comparison.}
CNN detectors (YOLO~\cite{yolov8_ultralytics}) and Transformer detectors (RT-DETR~\cite{lv2023detrs}, DINO~\cite{zhang2023dino}) exhibit distinct epistemic weight profiles when averaged over MOT17 and DanceTrack.
CNNs assign higher mass to cross layer instability (mean 0.60 vs.\ 0.40), whereas Transformers emphasize geometric collapse (0.42 vs.\ 0.26), consistent with global attention producing more structured feature manifolds.
Despite these shifts, both families maintain strong orthogonality ($|r| \approx 0.05$) and similar conformal behavior (coverage 89–90\%, interval width reduction 28–32\% vs.\ the confidence baseline), demonstrating robustness of the decomposition across architectural paradigms.

\section{Conclusion}

We introduced a unified framework for disentangled aleatoric and epistemic uncertainty in deep visual inference, together with a source-preserving conformal calibration method and an adaptive inference controller. The decomposition yields orthogonal components with stable behavior across datasets, model scales, and architectures, enabling calibrated prediction intervals and reliable uncertainty signals. Empirically, the framework improves ambiguity detection, selective prediction, and adaptive model selection while maintaining coverage guarantees and reducing compute. The results demonstrate that principled uncertainty factorization is both feasible and practically valuable for robust, resource-aware visual tracking. Future work will extend these ideas to multi-object reasoning, temporally aware conformal methods, and real-time deployment on edge platforms.

{
    \small
    \bibliographystyle{ieeenat_fullname}
    \bibliography{main}
}

\end{document}